\documentclass[runningheads]{llncs}
\usepackage{graphicx}

\usepackage{tikz}
\usepackage{comment} 
\usepackage{amsmath,amssymb} 
\usepackage{color}

\usepackage{caption}
\usepackage{wrapfig}
\usepackage{blindtext}
\usepackage{bbm}
\usepackage{enumitem}
\usepackage{epsfig}
\usepackage[ruled]{algorithm2e}
\usepackage{algorithmic}
\usepackage{multirow}
\usepackage{color}
\usepackage{xcolor}
\usepackage{color, colortbl}
\usepackage{array}
\usepackage[english]{babel} 
\usepackage{amsmath,amsfonts} 
\usepackage{mathtools}
\usepackage{pifont}

\usepackage[pagebackref=true,breaklinks=true,letterpaper=true,colorlinks,bookmarks=false]{hyperref}

\newcommand{\zs}{\textit{zero-shot}}

\newcommand{\seen}{\textit{seen}}
\newcommand{\unseen}{\textit{unseen}}
\newcommand{\x}{\mathbf{x}}
\newcommand{\z}{\mathbf{z}}
\newcommand{\vv}{\mathbf{v}}
\newcommand{\y}{y}
\newcommand{\cc}{\mathbf{c}}
\newcommand{\f}{f}

\newcommand{\s}{\mathbf{s}}
\newcommand{\ft}{\mathbf{t}}
\newcommand{\X}{\mathcal{X}}
\newcommand{\Z}{\mathcal{Z}}
\newcommand{\V}{\mathcal{V}}
\newcommand{\Y}{\mathcal{Y}}
\newcommand{\C}{\mathcal{C}}
\newcommand{\indep}{\perp \!\!\! \perp}
\newcommand{\eg}{\textit{e.g.}}
\newcommand{\ie}{\textit{i.e.}}
\newcommand{\wrt}{\textit{w.r.t.}}
\DeclareMathOperator*{\argmin}{arg\,min}
\DeclareMathOperator*{\argmax}{arg\,max}

\begin{document}
	\pagestyle{headings}
	\mainmatter
	\def\ECCVSubNumber{100}  
	
	\title{Invertible Zero-Shot Recognition Flows} 

	\titlerunning{Invertible Zero-Shot Recognition Flows}
	%
	\author{Yuming Shen\inst{1} \and
		Jie Qin\inst{2}\thanks{Corresponding author. This is not our final version to ECCV.} \and
		Lei Huang\inst{2}
	}
	\authorrunning{Y. Shen et al.}
	%
	\institute{eBay \and Inception Institute of Artificial Intelligence\\
		\email{ymcidence@gmail.com}}
	\maketitle
	
	\begin{abstract}
		Deep generative models have been successfully applied to Zero-Shot Learning (ZSL) recently. However, the underlying drawbacks of GANs and VAEs (\eg, the hardness of training with ZSL-oriented regularizers and the limited generation quality) hinder the existing generative ZSL models from fully bypassing the seen-unseen bias. To tackle the above limitations, for the first time, this work incorporates a new family of generative models (\ie, flow-based models) into ZSL. The proposed Invertible Zero-shot Flow (IZF) learns factorized data embeddings (\ie, the semantic factors and the non-semantic ones) with the forward pass of an invertible flow network, while the reverse pass generates data samples. This procedure theoretically extends conventional generative flows to a factorized conditional scheme. To explicitly solve the bias problem, our model enlarges the seen-unseen distributional discrepancy based on a negative sample-based distance measurement. Notably, IZF works flexibly with either a naive Bayesian classifier or a held-out trainable one for zero-shot recognition. Experiments on widely-adopted ZSL benchmarks demonstrate the significant performance gain of IZF over existing methods, in both classic and generalized settings.
		
	\end{abstract}
	
	\vspace{-2ex}\section{Introduction}\label{sec_1}\vspace{-1ex}
	
	With the explosive growth of image classes, there is an ever-increasing need for computer vision systems to recognize images from never-before-seen classes, a task which is known as Zero-Shot Learning (ZSL)~\cite{lampert2009learning}. Generally, ZSL aims at recognizing \unseen~images by exploiting relationships between \seen~and \unseen~images. Equipped with prior semantic knowledge (\eg, attributes~\cite{dap}, word embeddings~\cite{mikolov2013distributed}), traditional ZSL models typically mitigate the \seen-\unseen~domain gap by learning a visual-semantic projection between images and their semantics. In the context of deep learning \cite{shen2020exploiting,shen2019human}, the recent emergence of generative models has slightly changed this schema by converting ZSL into supervised learning, where a held-out classifier is trained for zero-shot recognition based on the generated \unseen~images. As both \seen~and synthesized \unseen~images are observable to the model, generative ZSL methods largely favor Generalized ZSL (GZSL)~\cite{gzsl} and yet perform well in Classic ZSL (CZSL)~\cite{lampert2009learning,thomas,wang2016zeroshot}.
	In practice, Generative Adversarial Networks (GANs)~\cite{gan}, Variational Auto-Encoders (VAEs)~\cite{vae} and Conditional VAEs (CVAEs)~\cite{cvae} are widely employed for ZSL. Despite the considerable success current generative models \cite{lisgan,cvaezsl,fclswgan,fvaegan,Zhu_2019_ICCV} have achieved, their underlying limitations are still inevitable in the context of ZSL.
	\begin{figure}[t]
		\begin{center}
			\includegraphics[width=.8\linewidth]{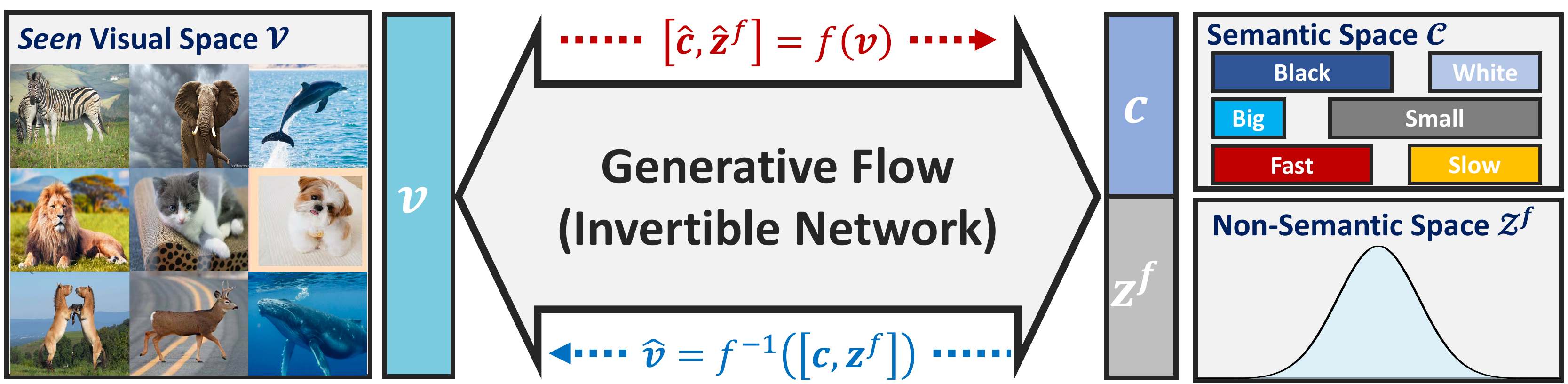}
		\end{center}\vspace{-2ex}
		\caption{A brief illustration of IZF for ZSL. We propose a novel factorized conditional \textbf{generative flow} with \text{invertible} networks.
		}\vspace{-3ex}
		\label{fig_0} 
	\end{figure}
	
	First, GANs~\cite{gan} suffer from mode collapse~\cite{modecollapse} and instability during training with complex learning objectives. It is usually hard to impose additional ZSL-oriented regularizers to the generative side of GANs other than the real/fake game \cite{cadavae}. Second, the Evidence Lower BOund (ELBO) of VAEs/CVAEs~\cite{vae,cvae} requires stochastic approximate optimization, preventing them from generating high-quality \unseen~samples for robust ZSL \cite{fvaegan}. Third, as only seen data are involved during training, most generative models are not well-addressing the \textbf{seen-unseen bias} problem, \ie, generated \unseen~data tend to have the same distribution as \seen~ones. Though these concerns are as well partially noticed by the recent ZSL research \cite{cadavae,fvaegan}, they either simply bypass the drawback of GAN in ZSL by resorting to VAE or \textit{vice versa}, which can be yet suboptimal.

	Therefore, we ought to seek a novel generative model that can bypass the above limitations to further boost the performance of ZSL.
	%
	%
	Inspired by the recently proposed Invertible Neural Networks (INNs)~\cite{inn}, we find that another branch of generative models, \ie, flow-based generative models~\cite{nice,realnvp}, align well with our insights into generative ZSL models. Particularly, generative flows adopt an identical set of parameters and built-in network for encoding (\textit{forward pass}) and decoding (\textit{reverse pass}). Compared with GANs/VAEs, the forward pass in flows acts as an additional `encoder' to fully utilize the semantic knowledge. Furthermore, flows can be easily extended into a conditional scheme to generate \unseen~data of good quality. 
	
	In this paper, we fully exploit the advantages of generative flows~\cite{nice,realnvp}, based on which a novel ZSL model is proposed, namely Invertible Zero-shot Flow (IZF). In particular, the forward pass of IZF projects visual features to the semantic embedding space, with the reverse pass consolidating the inverse projection between them. We adopt the idea of factorized representations in \cite{dlfzrl,mfm} to disentangle the output of the forward pass into two factors, \ie, semantic and non-semantic ones. Thus, it becomes possible to inject category-wise similarity knowledge into the model by regularizing the semantic factors. Meanwhile, the respective reverse pass of IZF performs conditional data generation with factorized embeddings for both \seen~and \unseen~data. We visualize this pipeline in Fig.~\ref{fig_0}. To further accommodate IZF to ZSL, we propose novel bidirectional training strategies to \textbf{(1)} centralize the \seen~prototypes for stable classification, and \textbf{(2)} diverge the distribution of synthesized \unseen~data and real \seen~data to explicitly address the bias problem. Our main contributions include:
	\begin{enumerate}\vspace{-0ex}
		\item IZF shapes a novel factorized conditional flow structure that supports exact density estimation. This differs from the existing approximated~\cite{inn} and the non-factorized~\cite{ardizzone2019guided} approach. To the best of our knowledge, IZF is the first generative flow model for ZSL.
		
		
		\vspace{-0ex}\item A novel mechanism tackling the bias problem is proposed with the merits of the generative nature of IZF, \ie, measuring and diversifying the sample-based \seen-\unseen~data distributional discrepancy.
		\vspace{-0ex}\item Extensive experiments on both real-world data and simulated data demonstrate the superiority of IZF over existing methods in terms of GZSL and CZSL settings. \vspace{-1ex}
	\end{enumerate}
	
	\vspace{-2ex}\section{Related Work}\vspace{-1ex}
	\noindent\textbf{Zero-Shot Learning.} ZSL~\cite{lampert2009learning} has been extensively studied in recent years. The evaluation of ZSL can be either classic (CZSL) or generalized (GZSL)~\cite{gzsl}, while recent research also explores the potential in retrieval~\cite{long2018towards,zsih}. CZSL excludes \seen~classes during test, while GZSL considers both \seen~and \unseen~classes, being more popular among recent articles~\cite{Cacheux_2019_ICCV,Elhoseiny_2019_ICCV,Jiang_2019_ICCV,Li_2019_ICCV}. To tackle the problem of \seen-\unseen~domain shift, 
	there propose three typical ways to inject semantic knowledge for ZSL, \ie, \textbf{(1)} learning visual$\rightarrow$semantic projections~\cite{ale,devise,sae,dap,eszsl}, \textbf{(2)} learning semantic$\rightarrow$visual projections~\cite{radovanovic2010hubs,dem,crnet}, and \textbf{(3)} learning shared features or multi-modal functions~\cite{sse}. Recently, deep generative models have been adapted to ZSL, subverting the traditional ZSL paradigm to some extent.
	The majority of existing generative methods employ GANs~\cite{lisgan,fclswgan,cewgan}, CVAEs~\cite{segzsl,cvaezsl,cadavae} or a mixture of the two~\cite{gdan,fvaegan} to synthesize \unseen~data points for a successive classification stage. 
	However, as mentioned in Sec.~\ref{sec_1}, these models suffer from their underlying drawbacks in the context of ZSL.
	
	
	\noindent\textbf{Generative Flows.} Compared with GANs/VAEs, flow-based generative models~\cite{nice,realnvp,glow} have attracted less research attention in the past few years, probably because this family of models require special neural structures that are in principle invertible for encoding and generation. It was not until the first appearance of the \text{coupling layer} in NICE~\cite{nice} and RealNVP~\cite{realnvp} that generative flows with deep INNs became practical and efficient. In \cite{Liu_2019_CVPR}, flows are extended to a conditional scheme, but the density estimation is not deterministic. The Glow architecture~\cite{glow} is further introduced with invertible 1$\times$1 convolution for realistic image generation. In~\cite{ardizzone2019guided}, conditions are injected into the coupling layers. IDF~\cite{idf} and BipartiteFlow~\cite{tran2019discrete} define a discrete case of flows. Flows can be combined with adversarial training strategies~\cite{flowgan}. In~\cite{waveglow}, generative flows have also been successfully applied to speech synthesis.
	
	\noindent \textbf{Literally Invertible ZSL.} We also notice that some existing ZSL models involve literally \textit{invertible} projections \cite{sae,zskl}. However, these methods are unable to generate samples, failing to benefit GZSL with the held-out classifier schema \cite{fclswgan} and our inverse training objectives. In addition, \cite{sae,zskl} are linear models and cannot be paralleled as deep neural networks during training. This limits their model capacity and training efficiency on large-scale data.
	
	\section{Preliminaries: Generative Flows and INNs}\label{sec_3}\vspace{-1ex}

	\noindent\textbf{Density Estimation with Flows.} Generative flows are theoretically based on the \textit{change of variables formula}. 
	Given a $d$-dimentional datum $\x\in\X\subseteq\mathbb{R}^d$ and a pre-defined prior $p_\Z$ supporting a set of latents $\z\in\Z\subseteq\mathbb{R}^d$, the \textit{change of variables formula} defines the estimated density of $p_\theta(\x)$ using an invertible (also called \textit{bijective}) transformation $f:\X\rightarrow\Z$ as follows:
	\vspace{-0ex}\begin{equation}\label{eq_1}
	p_\theta(\x)=p_\Z\left(\f\left(\x\right)\right)\left|\text{det}\frac{\partial \f}{\partial\x}\right|.
	\vspace{-0ex}\end{equation}
	Here $\theta$ indicates the set of model parameters and the scalar $\left|\text{det}\left(\partial \f/ \partial\x\right)\right|$ is the absolute value of the determinant of the Jacobian matrix $\left(\partial \f/ \partial\x\right)$. One can refer to \cite{nice,realnvp} and our \textbf{supplementary material} for more details. 
	The choice of the prior $p_\Z$ is arbitrary and a zero-mean unite-variance Gaussian is usually adequate, \ie, $p_\Z(\z)=\mathcal{N}(\z|\mathbf{0}, \mathbf{I})$. The respective generative process can be written as $\hat{\x} = \f^{-1}\left(\z\right), \text{where~} \z\sim p_\Z.$ 
	$\f$ is usually called the \textit{forward pass}, with $\f^{-1}$ being the \textit{reverse pass}.\footnote{Note that reverse pass and back-propagation are different concepts.} 
	Stacking a series of invertible functions $f=f_1\circ f_2\circ\cdots\circ f_k$ literally complies with the name of \textit{flows}. 
	
\begin{figure*}[t]
		\begin{center}
			\includegraphics[width=.95\textwidth]{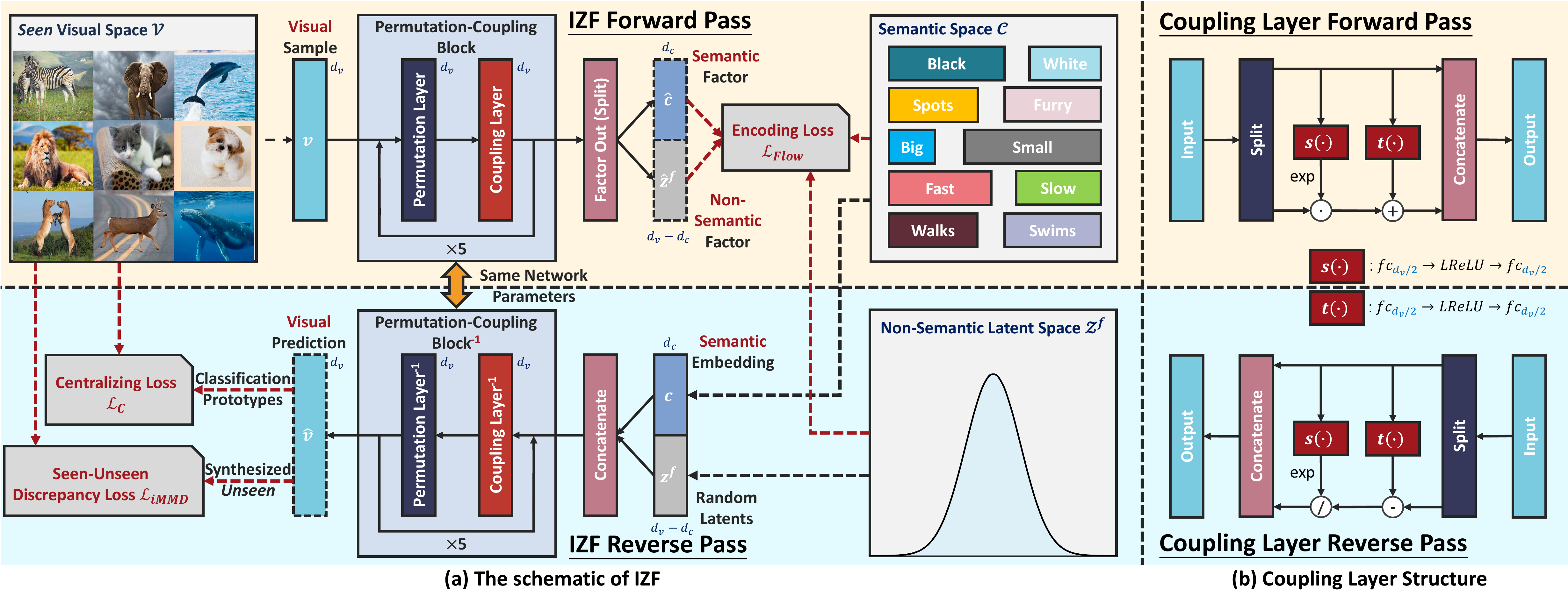}
		\end{center}\vspace{-3ex}
		\caption{\textbf{(a)} The architecture of the proposed IZF model. The forward pass and reverse pass are indeed sharing network parameters as invertible structures are used. Also note that only \seen~visual samples are accessible during training and IZF is an \textbf{inductive} ZSL model. \textbf{(b)} A typical illustration of the coupling layer~\cite{nice} used in our model.}\vspace{-2ex}
		\label{fig_2}
		
	\end{figure*}

	\noindent\textbf{INNs with Coupling Layers.} Generative flows admit networks with \textbf{(1)} exactly invertible structure and \textbf{(2)} efficiently computed Jacobian determinant. We adopt a typical type of INNs, called the coupling layers~\cite{nice}, which split network inputs/outputs into two respective partitions: $\x = \left[\x_a, \x_b\right]$, $\z = \left[\z_a, \z_b\right]$. The computation of the layer is defined as:
	\begin{equation}\label{eq_2}
	\begin{split}
	\f(\x)&=\left[\x_a, \x_b\odot\exp\left(\s(\x_a)\right) + \ft(\x_a)\right],\\
	\f^{-1}(\z)& = \left[\z_a,\left(\z_b-\ft(\z_a)\right)\oslash\exp\left(\s(\z_a)\right)\right],
	\end{split}
	\end{equation}
	where $\odot$ and $\oslash$ denote element-wise multiplication and division respectively. $\s(\cdot)$ and $\ft(\cdot)$ are two arbitrary neural networks with input and output lengths of $d/2$. We show this structure in Fig.~\ref{fig_2} (b). Its corresponding log-determinant of Jacobian can be conveniently computed by $\sum\left|\s\right|$. Coupling layers usually come together with element-wise permutation to build compact transformation.
	

	\vspace{-2ex}\section{Formulation: Factorized Conditional Flow}\label{sec_41}\vspace{-1ex}
	ZSL aims at recognizing \unseen~data. The training set $\mathcal{D}^{s}=\left\{(\vv^s, \y^s, \cc^s)\right\}$ of it is grounded on $M^s$ \seen~classes, \ie, {$\y^s\in\mathcal{Y}^s=\{1,2,...,M^s\}$}.  Let $\V^s\subseteq\mathbb{R}^{d_v}$ and $\C^s\subseteq\mathbb{R}^{d_c}$ respectively represent the visual space and the semantic space of \seen~data, of which $\vv^s\in\V^s$ and $\cc^s\in\C^s$ are the corresponding feature instances. The dimensions of these two spaces are denoted as $d_v$ and $d_c$. Given an \unseen~label set {$\Y^u=\{M^s+1, M^s+2,...,M^s+M^u\}$} of $M^u$ classes, the \unseen~data are denoted with the superscript of $\cdot^u$ as $\mathcal{D}^{u}=\left\{(\vv^u, \y^u, \cc^u)\right\}$, where $\vv^u\in\V^u$, $\y^u\in\Y^u$ and $\cc^u\in\C^u$. 
	In this paper, the superscript are omitted when the referred sample can be both \seen~or \unseen, \ie, $\vv\in\V=\V^s\cup\V^u,\y\in\Y=\Y^s\cup\Y^u$ and $\cc\in\C=\C^s\cup\C^u$.
	
	The framework of IZF is demonstrated in Fig.~\ref{fig_2} (a). IZF factors out the high-level semantic information with its forward pass $\f(\cdot)$, equivalently performing visual$\rightarrow$semantic projection. The reverse pass handles conditional generation, \ie, semantic$\rightarrow$visual projection, with identical network parameters to the forward pass. To reflect label information in a flow, Eq.~\eqref{eq_1} is slightly extended to a conditional scheme with visual data $\vv$ and their labels $\y$:
	\vspace{-1ex}\begin{equation}\label{eq_con}\vspace{-0ex}
	p_\theta(\vv|\y)=p_\Z\left(\f\left(\vv\right)|\y\right)\left|\text{det}\frac{\partial \f}{\partial\vv}\right|.
	\end{equation}
	Detailed proofs are given in the \textbf{supplementary material}. Next, we consider reflecting semantic knowledge in the encoder outputs for ZSL. To this end, a factorized model takes its shape.
	\vspace{-1ex}\subsection{Forward Pass: Factorizing the Semantics}\vspace{-1ex}
	
	High-dimensional image representations contain both high-level semantic-related information and non-semantic information such as low-level image details. As factorizing image features has been proved effective for ZSL in~\cite{dlfzrl}, we adopt this spirit, but with different approach to fit the structure of flow. In \cite{dlfzrl}, the factorization is basically only empirical, while IZF derives full likelihood model of a training sample. 
	
	As shown in Fig.~\ref{fig_2} (a), the proposed flow network learns factorized independent image representations $\hat{\z}=[\hat{\cc}, \hat{\z}^\f]=\f(\vv)$ with its forward pass $\f(\cdot)$, where $\hat{\cc}\in\mathbb{R}^{d_c}$ denotes the predicted semantic factor of an arbitrary visual sample $\vv$ and $\hat{\z}^\f\in\mathbb{R}^{d_v-d_c}$ is the low-level non-semantic independent to $\hat{\cc}$, \ie, $\hat{\z}^f\indep\hat{\cc}$. We assume $\hat{\z}^\f$ is not dependent on data label $y$, \ie, $\hat{\z}^f\indep \y$ as it is designed to reflect no high-level semantic/category information. 
	Therefore, we rewrite the conditional probability of Eq.~\eqref{eq_con} as
	\vspace{-1ex}\begin{equation}\label{eq_4}
	p_\theta(\vv|y)= p_\Z\big([\hat{\cc},\hat{\z}^f]=\f(\vv)|y\big)\left|\text{det}\frac{\partial\f}{\partial\vv}\right|
	=p_{\C|\Y}(\hat{\cc}|y)p_{\Z^f}(\hat{\z}^f)\left|\text{det}\frac{\partial\f}{\partial\vv}\right|.
	\vspace{-1ex}\end{equation}
	The conditional independence property gives $p_\Z(\hat{\cc},\hat{\z}^f|y)=p_{\C|\Y}(\hat{\cc}|y)p_{\Z^f}(\hat{\z}^f)$. According to \cite{betavae,mfm}, this property is implicitly enforced by imposing fix-formed priors on each variable. In this work, the factored priors are
	\vspace{-1ex}\begin{equation}\vspace{-1ex}
	\begin{split}
	p_{\C|\Y}(\hat{\cc}|\y) = \mathcal{N}(\hat{\cc}|\cc(y),\mathbf{I}),
	\quad p_{\Z^f}(\hat{\z}^f) = \mathcal{N}(\hat{\z}^f|\mathbf{0},\mathbf{I}),
	\end{split}
	\vspace{-1ex}\end{equation}
	where $\cc(\y)$ simply denotes the semantic embedding corresponding to $\y$. Similar to the likelihood computation of VAEs~\cite{vae}, we empirically assign a uniformed Gaussian to $p_{\C|\Y}(\hat{\cc}|\y)$ centered at the corresponding semantic embedding $\cc(y)$ of the visual sample so that it can be simply reduced to a $l2$ norm.\
	
	The conditional schema of Eq.~\eqref{eq_4} is different from the one of \cite{Liu_2019_CVPR} where an additional condition encoder is required. IZF involves no auxiliary conditional component by learning factorized latents.
	
	\noindent \textbf{The Injected Semantic Knowledge.} The benefits of the factorized $p_{\C|\Y}(\hat{\cc}|\y)$ are two-fold: \textbf{1)} it explicitly reflects the degree of similarity between different classes, ensuring smooth \seen-\unseen~generalization for ZSL. This is also in line with the main motivation of several existing approaches~\cite{sae,eszsl}; \textbf{2)} a well-trained IZF model with $p_{\C|\Y}(\hat{\cc}|\y)$ factorizes the semantic meaning from non-semantic information of an image, making it possible to conditionally generate samples with $\f^{-1}(\cdot)$ by directly feeding the semantic category embedding (see Eq.~\eqref{eq_5}). 
	\vspace{-2ex}\subsection{Reverse Pass: Conditional Sample Generation}
	\vspace{-1ex}One advantage of deep generative ZSL models is the ability to observe synthesized \unseen~data.
	IZF fulfills this by
	\vspace{-0ex}\begin{equation}\label{eq_5}
	\cc\in\C, \z^f\sim p_{\Z^f},\hat{\vv}=\f^{-1}\left([\cc,\z^f]\right).
	\end{equation}
	
	
	
	\noindent\textbf{The Use of Reverse Pass.} Different from most generative ZSL approaches \cite{cvaezsl,fclswgan} where synthesized \unseen~samples simply feed a held-out classifier, IZF additionally uses these synthesized samples to measure the biased distributional overlap between \seen~and synthesized \unseen~data. We will elaborate the corresponding learning objectives and ideas in Sec. \ref{sec_424}.
	
	\vspace{-2ex}\subsection{Network Structure}\label{sec_412}\vspace{-1ex}
	In the spirits of Eq.~\eqref{eq_4} and \eqref{eq_5}, we build the network of IZF as shown in Fig.~\ref{fig_2} (a). Concretely, IZF consists of 5 permutation-coupling blocks to shape a deep non-linear architecture. Inspired by~\cite{inn,realnvp}, we combine the coupling layer with channel-wise permutation in each block. The permutation layer shuffles the elements of an input feature in a random but fixed manner so that the split of two successive coupling layers are different and the encoding/decoding performance is assured. We use identical structure for the built-in neural network $\s(\cdot)\text{ and }\ft(\cdot)$ of the coupling layers in Eq.~\eqref{eq_2}, \ie, $\mathtt{fc}_{d_v/2}\rightarrow \mathtt{LReLU}\rightarrow\mathtt{fc}_{d_v/2}$, where $\mathtt{LReLU}$ is the leaky ReLU activation~\cite{lrelu}. 
	In the following, we show how the network is trained to enhance ZSL.
	
	\vspace{-1ex}\section{Training with the Merits of Generative Flow}\vspace{-1ex}
	
	To transfer knowledge from \seen~concepts to \unseen~ones, we employ the idea of bi-directional training of INNs~\cite{inn} to optimize IZF. In principle, generative flows can be trained only with the forward pass (Sec.~\ref{sec_421}). However, considering the fact that the reverse pass of IZF is used for \zs~classification, we impose additional learning objectives to its reverse pass to promote the ability of \seen-\unseen~generalization (Sec.~\ref{sec_423} and \ref{sec_424}).
	
	\vspace{0ex}\subsection{Learning to Decode by Encoding}\label{sec_421}\vspace{-1ex}
	
	The first learning objective of IZF comes from the definition of generative flow as depicted in Eq.~\eqref{eq_1}. By analytic log-likelihood maximization of the forward pass, generative flows are ready to synthesize data samples. As only visual features of \seen~categories are observable to IZF, we construct this loss term upon $\mathcal{D}^s$ as
	\vspace{-0ex}\begin{equation}
	\mathcal{L}_\text{Flow} = \mathbb{E}_{(\vv^s,\y^s,\cc^s)}\left[-\log p_\theta(\vv^s|\y^s)\right],
	\vspace{-0ex}\end{equation}
	where $(\vv^s,\y^s,\cc^s)$ are \seen~samples from the training set $\mathcal{D}^s$ and $p_\theta(\vv^s|\y^s)$ is computed according to Eq.~\eqref{eq_4}. $\mathcal{L}_\text{Flow}$ is not only an encoding loss, but also can legitimate unconditional \seen~data generation due to the invertible nature of IZF. Compared with the training process of GAN/VAE-based ZSL models~\cite{cvaezsl,fclswgan}, IZF defines an explicit and simpler objective to fulfill the same functionality.
	%
	\vspace{-2ex}\subsection{Centralizing Classification Prototypes}\label{sec_423}\vspace{-1ex}
	
	\begin{wrapfigure}{RT}{.5\linewidth}\vspace{-2ex}
		
		\begin{center}
			\includegraphics[width=\linewidth]{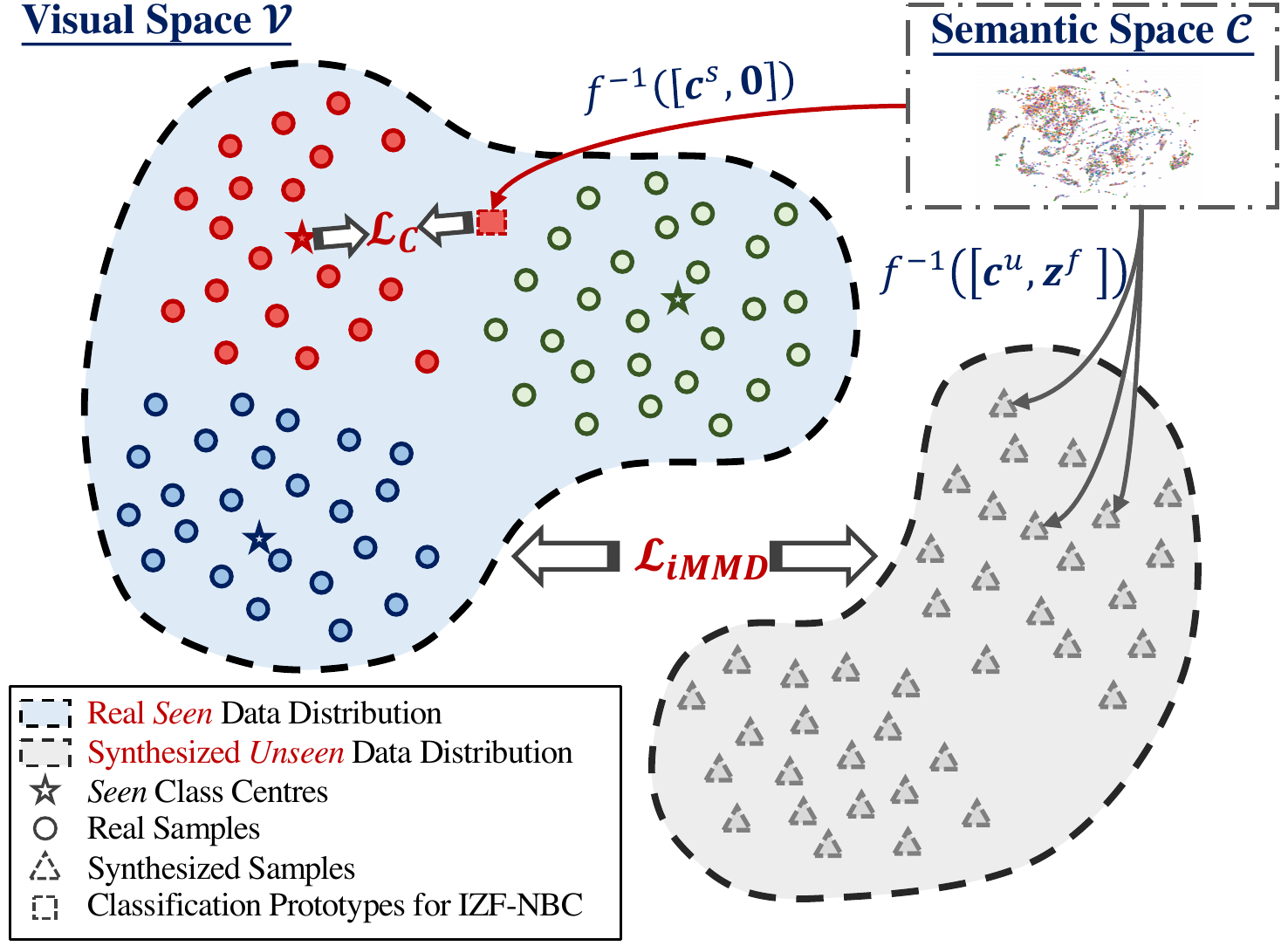}
		\end{center}\vspace{-2ex}
		\caption{Typical illustration of the IZF training losses \wrt~the \textbf{reverse pass}. In particular, $\mathcal{L}_\text{C}$ refers to the centralizing loss (Sec.~\ref{sec_423}) 
			for naive Bayesian classification. $\mathcal{L}_\text{iMMD}$ pushes the synthesized \unseen~visual distribution $p_{\hat{\V}^u}$ from colliding with the real \seen~one $p_{\V^s}$ to tackle the bias problem (Sec.~\ref{sec_424}).}\vspace{-2ex}
		\label{fig_loss} 
		
	\end{wrapfigure}
	
	IZF supports naive Bayesian classification by projecting semantic embeddings back to the visual space with its reverse pass. For each class-wise semantic representation, we define a special generation procedure $\hat{\vv}_c=\f^{-1}([\cc,\mathbf{0}])$ as the \textbf{classification prototype} of a class. As these prototypes are directly used to classify images by distance comparison, it would be harmful to the final accuracy when the prototypes are too close to unrelated visual samples. To address this issue, $\f^{-1}$ is expected to position them close to the centres $\bar{\vv}_c$ of the respective classes they belong to. This idea is illustrated in Fig.~\ref{fig_loss}, denoted as $\mathcal{L}_C$. In particular, this centralizing loss is imposed on the \seen~classes as
	\vspace{-0ex}\begin{equation}\vspace{-0ex}
	\begin{split}
	\mathcal{L}_\text{C}=\mathbb{E}&_{(\cc^s,\bar{\vv}_c^s)}\left[\parallel\f^{-1}([\cc^s,\mathbf{0}])-\bar{\vv}_c^s\parallel^2\right],\\
	\end{split}\vspace{-0ex}
	\end{equation}
	where $\bar{\vv}_c^s$ is the corresponding numerical mean of the visual samples that belong to the class with the semantic embedded $\cc^s$. Similar to the semantic knowledge loss, we directly apply $l2$ norm to the model to regularize its behavior. 
	
	\vspace{-2ex}\subsection{Measuring the \textit{Seen-Unseen} Bias}\label{sec_424}\vspace{-1ex}
	Recalling the \text{bias problem} in ZSL with generative models, the synthesized \unseen~samples could be unexpectedly too close to the real \seen~ones. This would significantly decrease the classification performance for \unseen~classes, especially in the context of GZSL where \seen~and \unseen~data are both available. We propose to explicitly tackle the bias problem by preventing the \textbf{synthesized \unseen} visual distribution $p_{\hat{\V}^u}$ from colliding with the \textbf{real \seen} one $p_{\V^s}$. In other words, $p_{\V^s}$ is slightly pushed away from $p_{\hat{\V}^u}$.
	
	Our key idea is illustrated in Fig.~\ref{fig_loss}, denoted as $\mathcal{L}_\text{iMMD}$. 
	With generative models, it is always possible to measure distributional discrepancy without acknowledging the true distribution parameters of $p_{\hat{\V}^u}$ and $p_{\V^s}$ by treating this as a negative two-sample-test problem. Hence, we resort to Maximum Mean
	Discrepancy (MMD)~\cite{inn,wae} as the measurement. Since we aim to increase the discrepancy, the last loss term of IZF is defined upon the \textbf{numerical negation} of $\operatorname{MMD}\left(p_{\V^s}|| p_{\hat{\V}^u}\right)$ in a batch-wise fashion as
	\vspace{-1ex}\begin{equation}\label{eq_mmd}\vspace{-0ex}
	\begin{split}
	\mathcal{L}_{\text{iMMD}}=&-\operatorname{MMD}\left(p_{\V^s}|| p_{\hat{\V}^u}\right)=\tfrac{2}{n^2}\sum_{i,j}\kappa(\vv^s_i,\hat{\vv}^u_j)\\\vspace{-1ex}
	&-\tfrac{1}{n(n-1)}\sum_{i\neq j}\left(\kappa(\vv^s_i,\vv^s_j) + \kappa(\hat{\vv}^u_i,\hat{\vv}^u_j)\right),\\\vspace{-1ex}
	\text{where~}& \vv^s_i\in\V^s,~\cc^u_i\in\C^u,~\z_i^\f\sim p_{\Z^\f},~\hat{\vv}^u_i=\f^{-1}([\cc^u_i, \z_i^\f]).
	\end{split}
	\end{equation}
	Here $n$ refers to the training batch size, and $\kappa(\cdot)$ is an arbitrary positive-definite reproducing kernel function. Importantly, as only \seen~visual samples $\vv^s_i$ are directly used and $\hat{\vv}^u_i$ are synthesized, $\mathcal{L}_{\text{iMMD}}$ is indeed an \textbf{inductive} objective. The same setting has also been adopted in recent inductive ZSL methods~\cite{dcn,cadavae,cvae,fclswgan}, \ie, the names of the \unseen~classes are accessible during training while their visual samples remain inaccessible. 
	We also note that replacing $\mathcal{L}_\text{iMMD}$ by simply tuning the values of \unseen~classification templates $\f^{-1}([\cc^u, \mathbf{0}])$ is infeasible in inductive ZSL since there exists no \unseen~visual reference sample for direct regularization.
	
	\noindent \textbf{Discussion: the Negative MMD.} Positive MMD has been previously used in several ZSL articles such as ReViSE~\cite{revise}. However, \cite{revise} employs MMD to align the cross-modal latent space (minimizing $\operatorname{MMD}(\texttt{seen\_1}||\texttt{seen\_2})$), while $\mathcal{L}_\text{iMMD}$ here solves the bias problem by slightly pushing the generated $p_{\hat{\V}^u}$ away from $p_{\V^s}$ (slightly increasing $\operatorname{MMD}(\texttt{seen}||\texttt{gen\_unseen})$). 
	We resort to this solution for the bias problem as \unseen~samples are unavailable in inductive ZSL. The possible side-effect of the large values of $\mathcal{L}_{\text{iMMD}}$ is also noticed which could confuse some generative models to produce unrealistic samples to favor the value of $\mathcal{L}_{\text{iMMD}}$. 
	
	\vspace{-2ex}\subsection{Overall Objective and Training}\vspace{-1ex}
	By combining the above-discussed losses, the overall learning objective of IZF can be simply written as
	\begin{equation}\label{eq_lossall}
	\mathcal{L}_\text{IZF}=\lambda_1\mathcal{L}_\text{Flow} + \lambda_2\mathcal{L}_\text{C} + \lambda_3\mathcal{L}_\text{iMMD}.
	\end{equation}
	Three hyper-parameters $\lambda_1, \lambda_2\text{~and~}\lambda_3$ are introduced to balance the contributions of different loss terms.
	IZF is fully differentiable \wrt~$\mathcal{L}_\text{IZF}$. Hence, the corresponding network parameters can be directly optimized with Stochastic Gradient Descent (SGD) algorithms.
	
	\vspace{-2ex}\subsection{Zero-Shot Recognition with IZF}\vspace{-1ex}
	We adopt two ZSL classification strategies (\ie, IZF-NBC and IZF-Softmax) that work with IZF. Specifically, IZF-NBC employs a naive Bayesian classifier to recognize a given test visual sample $\vv_q$ by comparing the Euclidean distances between it and the classification prototypes introduced in Sec~\ref{sec_423}. IZF-Softmax leverages a held-out classifier similar to the one used in \cite{fclswgan}. The classification processes are performed as
	\vspace{-0ex}\begin{equation}\label{eq_test}\vspace{-1ex}
	\begin{split}
	&\text{IZF-NBC: }\hat{\y}^q = \argmin_y \parallel\f^{-1}([\cc(y),\mathbf{0}])-\vv^q\parallel,\\
	&\text{IZF-Softmax: }\hat{\y}^q = \argmax_y \mathtt{softmax}\left(\mathtt{NN}(\vv^q)\right).
	\end{split}
	\end{equation}
	Here $\mathtt{NN}(\cdot)$ is a single-layered fully-connected network trained with generated \unseen~data and the {softmax} cross-entropy loss on top of the softmax activation. We use $\cc(y)$ to indicate the corresponding class-level semantic embedding of $y$ for convenience. Note that $y\in\Y^u$ in CZSL and $y\in\Y^s\cup\Y^u$ in GZSL. 
	
	
	\vspace{-1ex}\section{Experiments}

	\subsection{Implementation Details}
	IZF is implemented with the popular deep learning toolbox PyTorch~\cite{torch}. We build the INNs according to the framework of FrEIA~\cite{inn,ardizzone2019guided}. The network architecture is elaborated in Sec.~\ref{sec_412}. 
	The built-in networks $\s(\cdot)\text{ and }\ft(\cdot)$ of all coupling layers of IZF are shaped by $\mathtt{fc}_{d_v/2}\rightarrow \mathtt{LReLU}\rightarrow\mathtt{fc}_{d_v/2}$. Following~\cite{inn,wae}, we employ the Inverse Multiquadratic (IM) kernel $\kappa(\vv,\vv')=2d_v/\left(2d_v+\parallel\vv-\vv'\parallel^2\right)$ in Eq.~\eqref{eq_mmd} for best performance. We testify the choice of $\lambda_1, \lambda_2\text{~and~}\lambda_3$ within $\{0.1,0.5,1,1.5,2\}$ and report the results of $\lambda_1=2, \lambda_2=1,\lambda_3=0.1$ for all comparisons. The Adam optimizer~\cite{adam} is used to train IZF with a learning rate of $5\times 10^{-4}$ \wrt~$\mathcal{L}_\text{IZF}$. The batch size is fixed to 256 for all experiments.
	
	\subsection{Toy Experiments: Illustrative Analysis}\label{sec_toy}\vspace{-1ex}
	\begin{figure*}[t]
		\begin{center}
			\includegraphics[width=\textwidth]{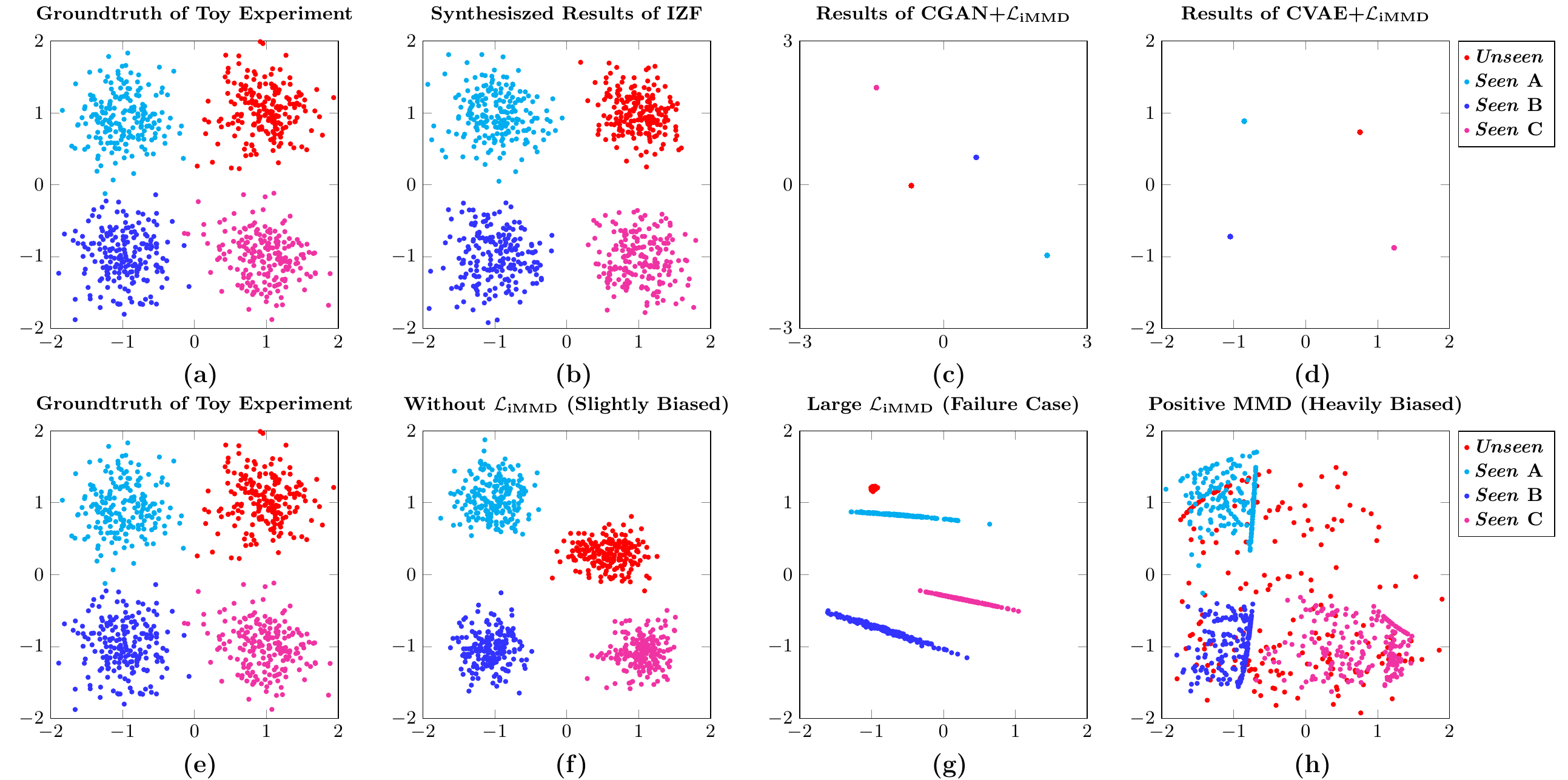}
		\end{center}\vspace{-3ex}
		\caption{Illustration of the 4-class toy experiment in Sec.~\ref{sec_toy}. \textbf{(a, e)} 2-D Ground truth simulation data, with the top-right class being \unseen. \textbf{(b)} Synthesized samples of IZF. \textbf{(c, d)} Synthesized results of conditional GAN and CVAE respectively with $\mathcal{L}_\text{iMMD}$. \textbf{(f)} Results without $\mathcal{L}_\text{iMMD}$ of IZF. \textbf{(g)} Failure results with extremely and unreasonably large $\mathcal{L}_\text{iMMD}$ ($\lambda_3=10$) of IZF. \textbf{(h)} Results with positive MMD of IZF.}
		\label{fig_toy1}\vspace{-3ex}
	\end{figure*}
	
	Before evaluating IZF with real data, we firstly provide a toy ZSL experiment to justify our motivation. Particularly, the following themes are discussed:
	\begin{enumerate}\vspace{-1ex}
		\item Why Do We Resort to Flows Instead of GAN/VAE with $\mathcal{L}_\text{iMMD}$?
		\item The effect of $\mathcal{L}_\text{iMMD}$ regarding the bias problem.\vspace{-1ex}
	\end{enumerate}
	
	\noindent \textbf{Setup.} We consider a 4-class simulation dataset with 1 class being \unseen. The class-wise attributes are defined as $\C^s=\{[0,1],[0,0],[1,0]\}$ for the \seen~classes \textbf{A}, \textbf{B} and \textbf{C} respectively, while the \unseen~class would have attribute of $\C^u=\{[1,1]\}$. The ground truth data are randomly sampled around a linear transformation of the attributes, \ie, $\vv \coloneqq 2\cc-1 + \epsilon\in\mathbb{R}^2$, where $\epsilon\sim\mathcal{N}(\mathbf{0}, \tfrac{1}{3}\mathbf{I})$. To meet the dimensionality requirement, \ie, $d_v>d_c$, we follow the convention of \cite{inn} to pad two zeros to data when feeding them to the network, \ie, $\vv'\coloneqq [\vv,0,0]$. The toy data are plotted in Fig.~\ref{fig_toy1} (a) and (e). 
	
	\noindent \textbf{Why Do We Resort to Flows Instead of GAN/VAE?} We firstly show the synthesized results of IZF in Fig.~\ref{fig_toy1} (b). It can be observed that IZF successfully interprets the relations of the \unseen~class to the \seen~ones, \ie, being closer to \textbf{A} and \textbf{C} but further to \textbf{B}. To legit the use of generative flow, we accordingly build two baselines by combining Conditional GAN (CGAN) and CVAE with our $\mathcal{L}_\text{iMMD}$ loss (see our \textbf{supplementary document} for implementation details). The respective generated results are shown in Fig.~\ref{fig_toy1} (c) and (d). Aligning with our motivation, $\mathcal{L}_\text{iMMD}$ quickly fails the unstable training process of GAN in ZSL. Besides, CVAE+$\mathcal{L}_\text{iMMD}$ isn't producing good-quality samples, undergoing the risk of obtaining biased classification hyper-planes of the held-out classifier. This is because the side-effects of $\mathcal{L}_\text{iMMD}$ would slightly skew the generated data distributions from being realistic with its negative MMD, which aggravates the drawbacks of unstable training (GAN) and inaccurate ELBO (VAE) discussed in Sec.~\ref{sec_1}. However, the stable-training and exact-likelihood-estimation properties of flows allow IZF to bypass the side-effects of $\mathcal{L}_\text{iMMD}$, fully utilizing it towards the \seen-\unseen~bias in ZSL.
	
	\begin{table*}[t]
		\centering
		\small
		\resizebox{\textwidth}{!}{
			\begin{tabular}{l lccc  ccc  ccc  ccc  ccc}
				\hline
				\rowcolor{gray!15}&&\multicolumn{3}{c}{\textbf{ AwA1~\cite{dap}}}&\multicolumn{3}{c}{\textbf{AwA2~\cite{dap}}}&\multicolumn{3}{c}{\textbf{CUB~\cite{cub}}}&\multicolumn{3}{c}{\textbf{SUN~\cite{sun}}}&\multicolumn{3}{c}{\textbf{aPY~\cite{apy}}}\\\hline
				\rowcolor{gray!15}{\textbf{Method}}& \textbf{Reference}& $A^s$ & $A^u$ & $H$ & $A^s$ & $A^u$ & $H$ & $A^s$ & $A^u$ & $H$ & $A^s$ & $A^u$ & $H$ & $A^s$ & $A^u$ & $H$\\ \hline\hline
				DAP~\cite{dap}& PAMI13& 88.7& 0.0& 0.0& 84.7& 0.0& 0.0& 67.9& 0.0& 0.0& 25.1& 4.2& 7.2&78.3& 4.8& 9.0\\
				CMT~\cite{cmt}&NIPS13& 86.9& 8.4& 15.3& 89.0& 8.7& 15.9& 60.1& 4.7& 8.7& 28.0& 8.7& 13.3& 74.2& 10.9& 19.0\\
				DeViSE~\cite{devise}&NIPS13& 68.7& 13.4& 22.4& 74.7& 17.1& 27.8& 53.0& 23.8& 32.8& 27.4& 16.9& 20.9& 76.9& 4.9& 9.2\\
				ALE~\cite{ale}&CVPR15& 16.8& 76.1& 27.5& 81.8& 14.0& 23.9& 62.8& 23.7& 34.4& 33.1& 21.8& 26.3& 73.7& 4.6& 8.7\\
				SSE~\cite{sse}&ICCV15& 80.5& 7.0& 12.9& 82.5& 8.1& 14.8& 46.9& 8.5& 14.4& 36.4& 2.1& 4.0& 78.9& 0.2& 0.4\\
				ESZSL~\cite{eszsl}&ICML15& 75.6& 6.6& 12.1& 77.8& 5.9& 11.0& 63.8& 12.6& 21.0& 27.9& 11.0& 15.8& 70.1& 2.4& 4.6\\
				LATEM~\cite{latem}&CVPR16& 71.1& 7.3& 13.3& 77.3& 11.5& 20.0& 57.3& 15.2& 24.0& 28.8& 14.7& 19.5& 73.0& 0.1& 0.2\\
				SAE~\cite{sae}&CVPR17& 77.1& 1.8& 3.5& 82.2& 1.1& 2.2& 54.0& 7.8& 13.6& 18.0& 8.8& 11.8& \textbf{80.9}& 0.4& 0.9\\
				DEM~\cite{dem}&CVPR17& 84.7& 32.8& 47.3& 86.4& 30.5& 45.1& 57.9& 19.6& 29.2& 34.3& 20.5& 25.6& 11.1& \textbf{75.1}& 19.4\\
				RelationNet~\cite{relationnet}&CVPR18& \textbf{91.3}& 31.4& 46.7& \textbf{93.4}& 30.0& 45.3& 61.1& 38.1& 47.0& -& -& -& -& -& -\\
				DCN~\cite{dcn}&NIPS18& 84.2& 25.5& 39.1& -& -& -& 60.7& 28.4& 38.7& 37.0& 25.5& 30.2& 75.0& 14.2& 23.9\\
				CRNet~\cite{crnet}&ICML19& 74.7& 58.1& 65.4& 78.8& 52.6& 63.1& 56.8& 45.5& 50.5& 36.5& 34.1& 35.3& 68.4& 32.4& 44.0\\
				LFGAA~\cite{lfgaa}&ICCV19& -& -& -& 90.3& 50.0& 64.4& 79.6& 43.4& 56.2& 34.9& 20.8& 26.1& -& -& -\\
				\hline
				CVAE-ZSL~\cite{cvaezsl}&ECCVW18& -& -& 47.2& -& -& 51.2& -& -& 34.5& -& -& 26.7& -& -& -\\
				SE-GZSL~\cite{segzsl}&CVPR18& 67.8& 56.3& 61.5& 68.1& 58.3& 62.8& 53.3& 41.5& 46.7& 30.5& 40.9& 34.9& -& -& -\\
				f-CLSWGAN~\cite{fclswgan}&CVPR18& 61.4& 57.9& 59.6& -& -& -& 57.7& 43.7& 49.7& 36.6& 42.6& 39.4& -& -& -\\
				LisGAN~\cite{lisgan}&CVPR19& 76.3& 52.6& 62.3& -& -& -& 57.9& 46.5& 51.6& 37.8& 42.9& 40.2& -& -& -\\
				SGAL~\cite{sgal}&NIPS19& 75.7& 52.7& 62.2& 81.2& 55.1& 65.6& 44.7& 47.1& 45.9& 31.2& 42.9& 36.1& -& -& -\\
				CADA-VAE~\cite{cadavae}&CVPR19& 72.8& 57.3& 64.1& 75.0& 55.8& 63.9& 53.5& 51.6& 52.4& 35.7& 47.2& 40.6& -& -& -\\
				GDAN~\cite{gdan}&CVPR19& -& -& -& 67.5& 32.1& 43.5& 66.7& 39.3& 49.5& \textbf{89.9}& 38.1& \text{53.4}& 75.0& 30.4& 43.4\\
				DLFZRL~\cite{dlfzrl}&CVPR19& -& -& 61.2& -& -& 60.9& -& -& 51.9& -& -& 42.5& -& -& 38.5\\
				f-VAEGAN-D2~\cite{fvaegan}&CVPR19& 70.6& 57.6& 63.5& -& -& -& 60.1& 48.4& 53.6& 38.0& 45.1&  41.3& -& -& -\\
				\hline
				\rowcolor{gray!15}\textbf{IZF-NBC}& \textbf{Proposed}& 75.2& {57.8}& {65.4}& 76.0& {58.1}& {65.9}& 56.3&44.2& 49.5& 50.6&44.5& 47.4& 58.3& 39.8& {47.3}\\
				\rowcolor{gray!15}\textbf{IZF-Softmax}& \textbf{Proposed}& 80.5& \textbf{61.3}& \textbf{69.6}& 77.5& \textbf{60.6}& \textbf{68.0}& 68.0&\textbf{52.7}& \textbf{59.4}& 57.0&\textbf{52.7}& \textbf{54.8}& 60.5& 42.3& \textbf{49.8}\\
				\hline
			\end{tabular}
		}\vspace{1ex}
		\caption{Inductive GZSL performance of IZF and the state-of-the-art methods with the PS setting~\cite{ps}. $A^s$ and $A^u$ are per-class accuracy scores (\%) on \seen~and \unseen~test samples, and $H$ denotes their harmonic mean. 
		}\vspace{-5ex}
		\label{tab_gzsl}
	\end{table*}
	
	\noindent \textbf{Towards the Bias Problem with $\mathcal{L}_\text{iMMD}$.} We also illustrate the effects of $\mathcal{L}_\text{iMMD}$ with more baselines. It is shown in Fig.~\ref{fig_toy1} (f) that the model is biased by the \seen~classes without $\mathcal{L}_\text{iMMD}$ (also see \texttt{Baseline 4} of Sec.~\ref{sec_541}). The \unseen~generated samples are positioned closely to the \seen~ones. This would be harmful to the employed classifiers when there exist multiple \unseen~categories. Fig.~\ref{fig_toy1} (g) is a failure case with large \seen-\unseen~discrepancy loss, which dominates the optimization process and overfits the network to generate unreasonable samples. We also discuss this issue in hyper-parameter analysis (see Fig.~\ref{fig_cmp} (c)). 
	Fig.~\ref{fig_toy1} (h) describes an extreme situation when employing positive MMD to IZF (negative $\lambda_3$, \texttt{Baseline 5} of Sec.~\ref{sec_541}). The generated \unseen~samples are forced to fit the \seen~distribution and thus, the network is severely biased.

	\vspace{-2ex}\subsection{Real Data Experimental Settings}\vspace{-1ex}

	\noindent \textbf{Benchmark Datasets.} Five datasets are picked in our experiments. Animals with Attributes (AwA1)~\cite{dap} contains 30,475 images of 50 classes and 85 attributes, of which AwA2 is a slightly extended version with 37,322 images. Caltech-UCSD Birds-200-20 (CUB)~\cite{cub} carries 11,788 images from 200 kinds of birds with 312-attribute annotations. SUN Attribute (SUN)~\cite{sun} consists of 14,340 images from 717 categories, annotated with 102 attributes. aPascal-aYahoo (aPY)~\cite{apy} comes with 32 classes with 64 attributes, accounting 15,339 samples. We adopt the \textbf{PS} train-test setting~\cite{ps} for both CZSL and GZSL.

	\noindent \textbf{Representations.} All images $\vv$ are represented using the 2048-D ResNet-101~\cite{resnet} features and the semantic class embeddings $\cc$ are category-wise attribute vectors from \cite{xian2018zero,ps}. We pre-process the image features with min-max rescaling.
	
	\noindent \textbf{Evaluation Metric.} For GZSL, we adopt the top-1 average per-class accuracy for comparison. The per-class accuracy of \seen~classes is denoted as $A^s$, with $A^u$ the accuracy on \unseen~classes. The harmonic mean $H$ of $A^s$ and $A^u$ is reported as well. As to CZSL, the identical per-class accuracy is used as measurement.
	
	
	\vspace{-2ex}\subsection{Comparison with the State-of-the-Arts}\vspace{-1ex}
	
	\noindent\textbf{Baselines.} IZF is compared with the state-of-the-art ZSL methods, including DAP~\cite{dap}, CMT~\cite{cmt}, SSE~\cite{sse}, ESZSL~\cite{eszsl}, SAE~\cite{sae}, LATEM~\cite{latem}, ALE~\cite{ale}, DeViSE~\cite{devise}, DEM~\cite{dem},  RelationNet~\cite{relationnet}, DCN~\cite{dcn}, CVAE-ZSL~\cite{cvaezsl}, SE-GZSL~\cite{segzsl}, f-CLSWGAN~\cite{fclswgan}, CRNet~\cite{crnet}, LisGAN~\cite{lisgan}, SGAL~\cite{sgal}, CADA-VAE~\cite{cadavae}, GDAN \cite{gdan}, DLFZRL\cite{dlfzrl}, f-VAEGAN-D2~\cite{fvaegan} and LFGAA~\cite{lfgaa}. 
	We report the official results of these methods from referenced articles with the identical experimental setting used in this paper for fair comparison.
	
	\begin{wraptable}{RT}{.5\linewidth}\vspace{-3ex}
		\small
		\centering
		\vspace{-1ex}\resizebox{\linewidth}{!}{
			\begin{tabular}{l c c c c c}
				\hline
				\rowcolor{gray!15}\textbf{Method}&\multicolumn{1}{c}{\textbf{AwA1}}&\multicolumn{1}{c}{\textbf{AwA2}}&\multicolumn{1}{c}{\textbf{CUB}}&\multicolumn{1}{c}{\textbf{SUN}}&\multicolumn{1}{c}{\textbf{aPY}}\\\hline\hline
				DAP~\cite{dap}& 44.1& 46.1& 40.0& 39.9& 33.8\\
				CMT~\cite{sse}& 39.5& 37.9& 34.6& 39.9& 28.0\\
				SSE~\cite{sse}& 60.1& 61.0& 43.9& 51.5& 34.0\\
				ESZSL~\cite{eszsl}& 58.2& 58.6& 53.9& 54.5& 38.3\\
				SAE~\cite{sae}& 53.0& 54.1& 33.3& 40.3& 8.3\\
				LATEM~\cite{latem}& 55.1& 55.8& 49.3& 55.3& 35.2\\
				ALE~\cite{ale}& 59.9& 62.5& 54.9& 58.1& 39.7\\
				DeViSE~\cite{devise}& 54.2& 59.7& 52.0& 56.5& 39.8\\
				RelationNet~\cite{relationnet}& 68.2& 64.2& 55.6& -& -\\
				DCN~\cite{dcn}& 65.2& -& 56.2& 61.8& 43.6\\
				f-CLSWGAN~\cite{fclswgan}& 68.2& -& 57.3& 60.8& -\\
				LisGAN~\cite{lisgan}& 70.6& -& 58.8& 61.7& 43.1\\
				DLFZRL~\cite{dlfzrl}& 61.2& 60.9& 51.9& 42.5& 38.5\\
				f-VAEGAN-D2~\cite{fvaegan}& 71.1& -& 61.0& 65.6&-\\
				LFGAA~\cite{lfgaa}& -& 68.1& \textbf{67.6}& 62.0& -\\
				\hline
				\rowcolor{gray!15}\textbf{IZF-NBC}& {72.7}& {71.9}& {59.6}& {63.0}& \textbf{45.2}\\
				\rowcolor{gray!15}\textbf{IZF-Softmax}& \textbf{74.3}& \textbf{74.5}& {67.1}& \textbf{68.4}& \text{44.9}\\
				\hline
			\end{tabular}
		}\vspace{-2ex}
		\caption{CZSL per-class accuracy (\%) comparison with the \textbf{PS} setting~\cite{ps}.}\label{tab_czsl}\vspace{-4ex}
	\end{wraptable}
	
	\noindent \textbf{Results.} The GZSL comparison results are shown in Tab.~\ref{tab_gzsl}. It can be observed that deep generative models obtains better on-average ZSL scores than the non-generative ones, while some simple semantic-visual projecting models hit comparable accuracy to them such as CRNet~\cite{crnet}. IZF-Softmax generally outperforms the compared methods, where the performance margins on AwA~\cite{dap} are significant. 
	IZF-NBC also works well on AwA~\cite{dap}
	The proposed model produces balanced accuracy between \seen~and \unseen~data and obtains significant higher \unseen~accuracy. This shows the effectiveness of the discrepancy loss $\mathcal{L}_\text{iMMD}$ in solving the bias problem of ZSL. In addition to the GZSL results, we conduct CZSL experiments as well, which is shown in Tab.~\ref{tab_czsl}. As a relatively simpler setting, CZSL provides direct clues of the ability to transform knowledge from \seen~to \unseen. 

	
	\vspace{-1ex}\subsection{Component Analysis}\label{sec_541}\vspace{-1ex}
	
	We evaluate the effectiveness of each component of IZF to legitimate our design, including the loss terms and overall network structure. The following baselines are proposed. 
	\textbf{(1) $\text{CVAE} + \mathcal{L}_\text{C} + \mathcal{L}_\text{iMMD}$.} We firstly show the importance of generative flow for our task by replacing it with a simple CVAE~\cite{cvae} structure. This baseline uses the semantic representation as condition, and outputs synthesized visual features. In addition to the Evidence Lower BOund (ELBO) of CVAE, $\mathcal{L}_\text{C}$ and $\mathcal{L}_{iMMD}$ are applied to the baseline. 
	\textbf{(2) Without $\mathcal{L}_\text{C}$ \& $\mathcal{L}_\text{iMMD}$.} All regularization on the reverse pass is omitted. 
	\textbf{(3) Without $\mathcal{L}_\text{C}$.} The prototype centralizing loss is removed. 
	\textbf{(4) Without $\mathcal{L}_\text{iMMD}$.} The discrepancy loss to control the \seen-\unseen~bias problem of ZSL is deprecated. 
	\textbf{(5) Positive MMD.} In Eq.~\eqref{eq_mmd}, we employ negative MMD to tackle the bias problem. We propose a baseline with a positive MMD version of it to study its influence. This is realized by setting $\lambda_3=-1$. 
	\textbf{(6) IM Kernel$\rightarrow$Gaussian Kernel.} Instead of the Inverse Multiquadratic kernel, another widely-used kernel function, \ie, the Gaussian kernel, is tested in implementing Eq.~\eqref{eq_mmd}.
	\begin{table}[t]
		\begin{minipage}{.55\linewidth}\vspace{4ex}
			\resizebox{\textwidth}{!}{
				\begin{tabular}{ll ccc  ccc}
					\hline
					\rowcolor{gray!15}& & \multicolumn{3}{c|}{\textbf{NBC}}& \multicolumn{3}{c}{\textbf{Softmax}}\\\hline
					\rowcolor{gray!15}&\textbf{Baseline}& $A^s$& $A^u$& $H$ &$A^s$& $A^u$& $H$\\\hline\hline
					1&$\text{CVAE} + \mathcal{L}_\text{C} + \mathcal{L}_\text{iMMD}$ & 65.1& 30.8 & 41.8& 71.1& 36.8 & 48.5\\
					2&Without $\mathcal{L}_\text{C}$ and $\mathcal{L}_\text{iMMD}$ & 66.0 & 43.4 & 52.7& 78.9& 38.1& 51.4\\
					3&Without $\mathcal{L}_\text{C}$ & 67.0& 41.7& 51.4& 79.2& 60.9& 68.8\\
					4&Without $\mathcal{L}_\text{iMMD}$& 79.6& 49.0 & 60.7& 81.3& 53.2& 64.3\\
					5&Positive MMD & 76.2& 21.1& 33.0& 80.7& 44.5& 57.4\\
					6&IM Kernel$\rightarrow$Gaussian Kernel& 73.6& 54.9& 62.9& 79.6& 61.7& 69.5 \\\hline
					\rowcolor{gray!15}&\textbf{IZF (full model)}& 75.2& 57.8& 65.4& 80.5& 61.3& 69.6\\\hline
				\end{tabular}
			}
			\vspace{3.5ex}\caption{Component analysis results on AwA1~\cite{dap} (Sec.~\ref{sec_541}). \textbf{NBC}: results with distance-based classifier. \textbf{Softmax}: results with a held-out trainable classifier.}
			\label{tab_abl}
		\end{minipage}\hspace{1ex}
		\begin{minipage}{.43\linewidth}
			
			\centering
			\includegraphics[width=\textwidth]{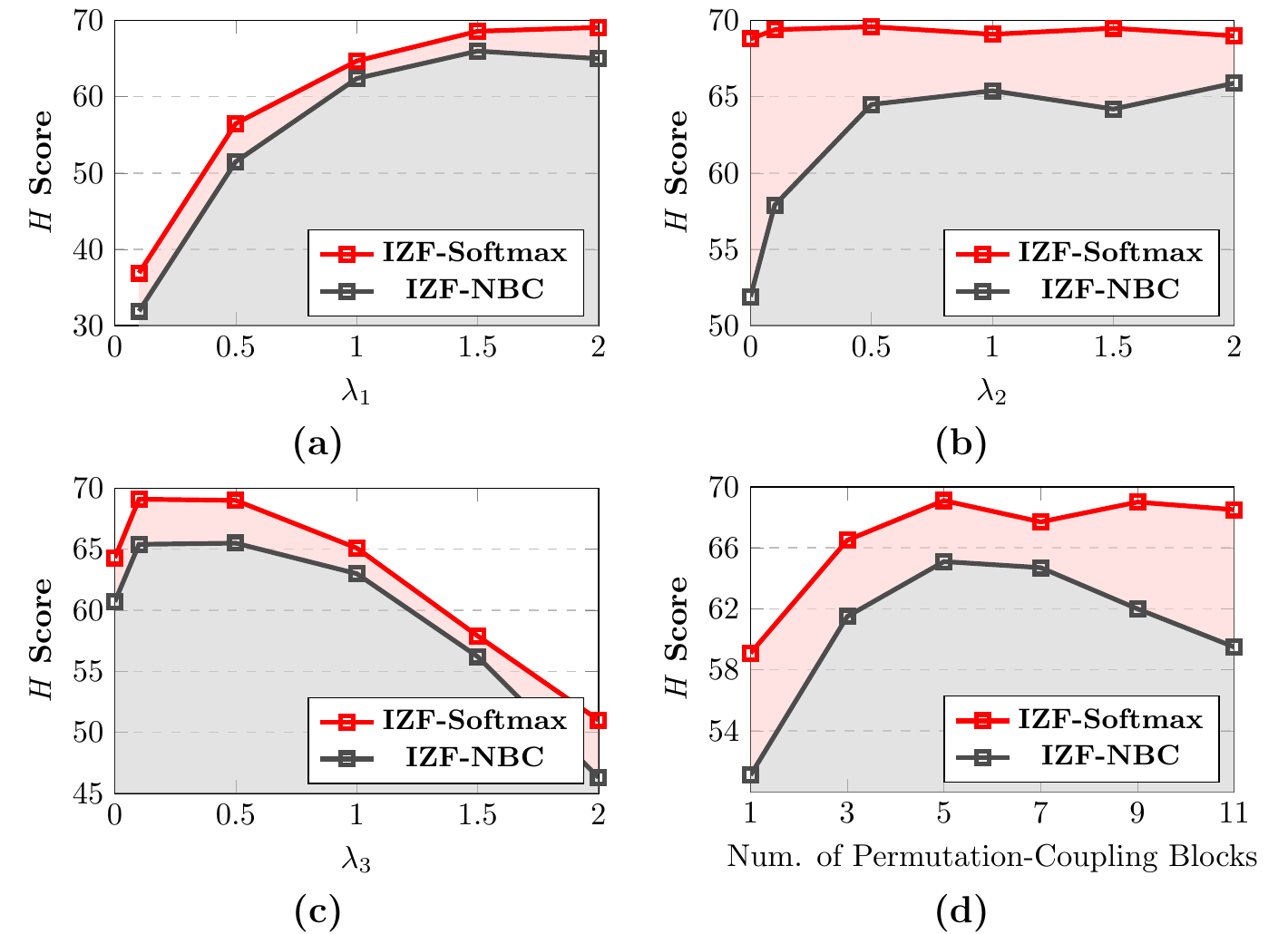}
			\captionof{figure}{\textbf{(a)}, \textbf{(b)} and \textbf{(c)} Hyper-parameter analysis for $\lambda_1$, $\lambda_2$ and $\lambda_3$. \textbf{(d)} Effect \wrt~numbers of the permutation-coupling blocks.}\vspace{-0ex}\label{fig_cmp}
		\end{minipage}
		
		\vspace{-5ex}
	\end{table}\vspace{-0ex}
	
	
	\noindent \textbf{Results.} The above-mentioned baselines are compared in Tab.~\ref{tab_abl} on AwA1~\cite{dap}. The GZSL criteria are adopted here as they are more illustrative metrics for IZF, showing different performance aspects of the model. Through our test, \texttt{Baseline 1}, \ie, $\text{CVAE} + \mathcal{L}_\text{C} + \mathcal{L}_\text{iMMD}$, is not working well with the distance-based classifier (Eq.~\eqref{eq_test}). 
	With loss components omitted (\texttt{Baseline 2-4}), IZF does not work as expected. In \texttt{Baseline 4}, the classification results are significantly biased to the \seen~concepts. 
	When imposing positive MMD to the loss function, the test accuracy of \seen~classes increases while the accuracy of \unseen~data drops quickly. This is because the bias problem gets severer and all generated samples, including the \unseen~classification prototypes, overfit to the \seen~domain. 
	The choice of kernel is not a key factor in IZF, and \texttt{Baseline 7} obtains on-par accuracy to IZF. Similar to GAN/VAE-based models~\cite{lisgan,cvaezsl,fclswgan}, IZF works with a held-out classifier, but it requires additional computational resources.
	
	\vspace{-2ex}\subsection{Hyper-Parameters}\label{sec_542}
	
	\vspace{-1ex}IZF involves 3 hyper-parameters in balancing the contribution of different loss items, shown in Eq.~\eqref{eq_lossall}. The influences of the values of them on AwA1 are plotted in Fig.~\ref{fig_cmp} (a), (b) and (c) respectively. A large weight is imposed to the semantic knowledge loss $\mathcal{L}_\text{Flow}$, \ie, $\lambda_1=2$, for best performance, as it plays an essential role in formulating the normalizing flow structure that ensures data generation with the sampled conditions and latents. A well-regressed visual-semantic projection necessitates conditional generation and, hence, bi-directional training. On the other hand, it is notable that a large value of $\lambda_3$ fails IZF overall. A heavy penalty to $\mathcal{L}_\text{iMMD}$ overfits the network to generate unreasonable samples to favour large \seen-\unseen~distributional discrepancy, and further prevents the encoding loss $\mathcal{L}_\text{Flow}$ from functioning. We observe significant increase of $\mathcal{L}_\text{Flow}$ throughout the training steps with $\lambda_3=2$, though $\mathcal{L}_\text{iMMD}$ decreases quickly. The performance of IZF \wrt~different numbers of permutation-coupling blocks is reported in Fig.~\ref{fig_cmp} (d), where we use 5 blocks in all comparisons. In general, IZF-Softmax is less sensitive to the depth of the network than IZF-NBC, but deeper networks would have higher likelihood to produce infinite gradients as coupling layers~\cite{realnvp} involve element-wise division. We further report the training efficiency of IZF in Fig.\ref{fig_tsne}~(c), where IZF only requires $\sim$20 epochs to obtain best-performing parameters.

	\vspace{-2ex}\subsection{Discriminability on \textit{Unseen} Classes}
	\begin{figure}[t]
		\begin{center}
			\includegraphics[width=\linewidth]{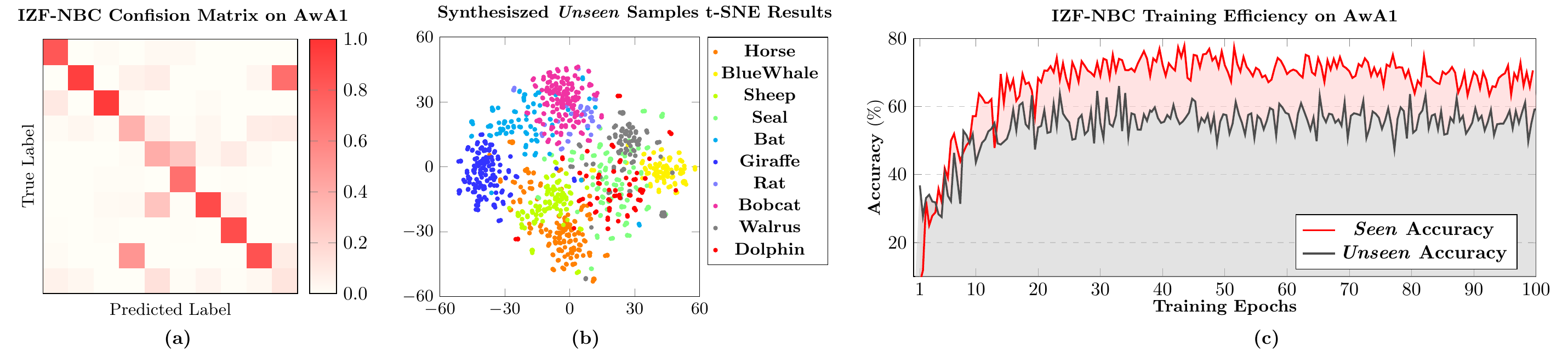}
		\end{center}\vspace{-4ex}
		\caption{\textbf{(a)} Confusion matrix of IZF on AwA1 with the CZSL setting. The order of labels is identical to the t-SNE legend. \textbf{(b)} t-SNE~\cite{tsne} results of the synthesized \unseen~samples on AwA1. \textbf{(c)} Training efficiency of IZF-NBC on AwA1. 
		}
		\label{fig_tsne}\vspace{-3ex}
	\end{figure}
	\vspace{-1ex}We intuitively analyze the discriminability and generation quality of IZF on \unseen~data by plotting the generated samples. The t-SNE~\cite{tsne} visualization of synthesized \unseen~data on AwA1~\cite{dap} is shown in Fig.~\ref{fig_tsne} (b). Although no direct regularization loss is applied to \unseen~classes, IZF manages to generate distinguishable samples according to their semantic meanings. In addition, the CZSL confusion matrix on AwA1 is reported in Fig.~\ref{fig_tsne} (a) as well.
	
	\vspace{-2ex}\section{Conclusion}\vspace{-2ex}
	In this paper, we proposed Invertible Zero-shot Flow (IZF), fully leveraging the merits of generative flows for ZSL. The invertible nature of flows enabled IZF to perform bi-directional mapping between the visual space and the semantic space with identical network parameters. The semantic information of a visual sample was factored-out with the forward pass of IZF. The classification prototypes of the reverse pass were regularized to stabilize distance-based classification. Last but not least, to handle the bias problem, IZF penalized \seen-\unseen~similarity by computing kernel-based distribution discrepancy with the generated data. The proposed model consistently outperformed state-of-the-art baselines on benchmark datasets.
	
	\clearpage
	%
	%
	\bibliographystyle{splncs04}
	\bibliography{egbib}

\end{document}